\documentclass[pdflatex,sn-mathphys-num]{sn-jnl}% Math and Physical Sciences Numbered Reference Style
\usepackage{lineno}
\usepackage{graphicx}%
\usepackage{multirow}%
\usepackage{amsmath,amssymb,amsfonts}%
\usepackage{amsthm}%
\usepackage{mathrsfs}%
\usepackage[title]{appendix}%
\usepackage{xcolor}%
\usepackage{textcomp}%
\usepackage{manyfoot}%
\usepackage{booktabs}%
\usepackage{algorithm}%
\usepackage{algorithmicx}%
\usepackage{algpseudocode}%
\usepackage{listings}%
\usepackage{setspace}
\usepackage{booktabs}
\usepackage{multirow}
\theoremstyle{thmstyleone}%
\theoremstyle{thmstyletwo}%

\theoremstyle{thmstylethree}%

\begin{document}

\title[Article Title]{Discovering Latent Response Laws in Forced Physical Systems}


\author[1]{\fnm{Yi} \sur{Zhu}}

\author*[1]{\fnm{Su} \sur{Chen}}\email{chensuchina@126.com}

\author[1]{\fnm{Xiaojun} \sur{Li}}

\author[1]{\fnm{Xiuli} \sur{Du}}

\affil*[1]{%
  \orgname{State Key Laboratory of Bridge Safety and Resilience, Beijing University of Technology},
  \orgaddress{%
    \city{Beijing},
    \postcode{100124},
    \country{China}
  }
}


\abstract{Governing equations provide compact descriptions of physical systems, yet the variables in which they are simple are often hidden in high-dimensional measurements. This challenge is sharper for forced systems, whose responses depend on both intrinsic dynamics and time-dependent inputs. Here we introduce FLARE, a forced latent autoencoder for response equations that learns compact response coordinates, identifies sparse input-dependent latent dynamics and decodes equation rollouts to full responses. By estimating latent dimension from data and separating state estimation from external forcing, FLARE enables forecasts to be initialized from past responses and driven by prescribed future inputs. Across known dynamical systems, application-scale forced responses and visual observations, FLARE recovers compact forced dynamics and predicts long-horizon high-dimensional responses under inputs not used for training. By turning learned coordinates into a dynamical interface, FLARE extends equation discovery to systems whose effective states are hidden within complex observations, providing a route for interpretable modelling and prediction of high-dimensional responses in forced dynamical systems.}

\keywords{equation discovery, forced systems, latent dynamics, response prediction}



\maketitle

\section{Introduction}\label{sec1}
Governing equations provide a compact language for describing, predicting and interrogating the behaviour of physical systems. When these equations are unknown or incomplete, data-driven discovery offers a complementary route to identifying effective models directly from observations. Symbolic regression and sparse system identification have demonstrated that both the structure and parameters of dynamical equations can be inferred from data across a broad range of systems governed by ordinary and partial differential equations \cite{Bongard2007AutomatedReverseEngineering,Schmidt2009DistillingNaturalLaws,Brunton2016SINDy,Rudy2017PDEFind,Schaeffer2017LearningPDEs,Chen2021PhysicsInformedScarceData,Long2019PDENet2,Both2021DeepMoD,Udrescu2020AIFeynman,Ruan2026ParallelSymbolicEnumeration}. Modern experiments and numerical simulations, however, do not always provide direct access to the variables required for a compact description of the dynamics. The dynamically relevant variables are often embedded in high-dimensional observations whose ambient dimension may greatly exceed the intrinsic dimensionality of the underlying dynamics \cite{Champion2019CoordinatesEquations,Chen2022FundamentalVariables,Lu2022PartialObservations,Bakarji2023DeepDelayAutoencoders}. At the same time, many systems are driven or controlled by time-dependent external inputs. For such systems, the scientific task extends beyond identifying evolution laws in known coordinates to discovering a compact representation of the response from high-dimensional observations and characterising how that response changes under external forcing \cite{Stark1997ForcedEmbedding,Proctor2016DMDc,Brunton2016SINDYc}.

Identifying low-dimensional dynamics from high-dimensional observations has motivated a broad range of reduced-order and latent-dynamics models. Dynamic mode decomposition and Koopman-based methods seek coordinates with simple linear evolution, whereas autoencoders, temporal models and neural operators use nonlinear representations to reconstruct and forecast complex dynamics \cite{Hinton2006DimensionalityReduction,Schmid2010DMD,Williams2015EDMD,Brunton2016KoopmanInvariantSubspaces,Lusch2018UniversalLinearEmbeddings,Otto2019LinearlyRecurrentAE,Vlachas2018HighDimensionalForecasting,Lu2021DeepONet,Li2021FourierNeuralOperator,Regazzoni2024LatentDynamicsNetworks,Kontolati2024LatentDeepONet,Gao2024GenerativeEffectiveDynamics}. These approaches have substantially advanced the reconstruction and prediction of high-dimensional systems. In parallel, equation-discovery methods seek suitable low-dimensional coordinates in which the system evolution admits an explicit and parsimonious representation. SINDy autoencoders introduced a framework for jointly learning such coordinates and sparse dynamics \cite{Champion2019CoordinatesEquations}; subsequent work extended this idea to equation discovery with uncertainty quantification \cite{Gao2024BayesianAutoencoders}, reduced modelling of high-dimensional flow fields \cite{Fukami2021LowDimensionalFlow}, and latent-dynamics identification from sparse measurements \cite{Gao2026SINDySHRED}. Related studies have sought to recover fundamental variables and governing relations directly from images and videos \cite{Chen2022FundamentalVariables,Luan2022VideoLaws,Li2026Pixel2Phys}. Together, these developments illustrate how representation learning and equation discovery can be combined to obtain a low-dimensional description in which the system evolution is expressed compactly. Within this framework, representation learning organizes complex observations into a small number of effective variables, whereas equation discovery characterises the relations governing their evolution \cite{Cranmer2020SymbolicModels}. Their combination therefore provides a natural route to reduced models that retain the essential features of high-dimensional responses while exposing the structure of the underlying dynamics.

Input-driven systems pose a distinct modelling problem. In such systems, the observed response depends not only on the intrinsic dynamics, but also on the external inputs and their coupling to the system state. Autonomous and forced models therefore address different scientific questions: the former describe the free evolution of a system, whereas the latter characterise a family of responses induced by different input histories. Dynamic mode decomposition with control, SINDy with control and Koopman-based model predictive control have established effective routes for incorporating external inputs into data-driven dynamical models \cite{Proctor2016DMDc,Brunton2016SINDYc,Kaiser2018SINDYMPC,Korda2018KoopmanMPC,Morton2018DeepDynamicalControl}. When the system response is high-dimensional, this problem intersects with representation learning. The learned low-dimensional representation must remain valid across forcing conditions, while the associated dynamics must capture how external inputs shape the response and remain predictive for input histories not encountered during training. In this setting, an explicit response equation provides a unified dynamical description of an input-driven system, enabling state--input interactions to be examined, behaviours across forcing conditions to be compared and responses to prescribed inputs to be predicted. However, when system responses are available only as high-dimensional observations, the effective variables required by the governing equations—and even the appropriate dimensionality of the latent space—are themselves unknown. The central challenge is therefore to identify directly from high-dimensional data both a compact response representation and its input-dependent evolution law, thereby linking external forcing, low-dimensional dynamics and the full system response within a single model \cite{Champion2019CoordinatesEquations,Brunton2016SINDYc,Bakarji2023DeepDelayAutoencoders,Lu2022PartialObservations,Gao2026SINDySHRED}.

To this end, we introduce the Forced Latent Autoencoder for Response Equations (FLARE), which jointly learns compact coordinates for high-dimensional responses and sparse input-dependent dynamics. FLARE estimates the intrinsic dimension of a latent point cloud to determine a compact latent space, encodes the latent state from response histories, explicitly incorporates external inputs into a sparse evolution equation, and decodes the resulting latent trajectory into the full response. In this way, response representation, external forcing and high-dimensional prediction are unified within a single dynamical model. Video sequences pose an additional representational challenge because the dynamical variables are embedded in pixel fields that evolve over time. To accommodate such observations, we replace the fully connected encoder and decoder with convolutional counterparts while retaining the input-dependent latent equation, thereby extending the same modelling principle from vector-valued observations to image sequences. We evaluate FLARE across a hierarchy of tasks spanning synthetic benchmarks, real-world applications and video observations. The results show that FLARE identifies compact forced dynamics and accurately predicts high-dimensional responses induced by inputs not encountered during training. FLARE thus provides a unified framework for discovering input-dependent dynamics from complex observations and using the resulting models for full-state prediction.

\section{Results}\label{sec2}

\begin{figure}[t!]
\centering
\includegraphics[width=\textwidth]{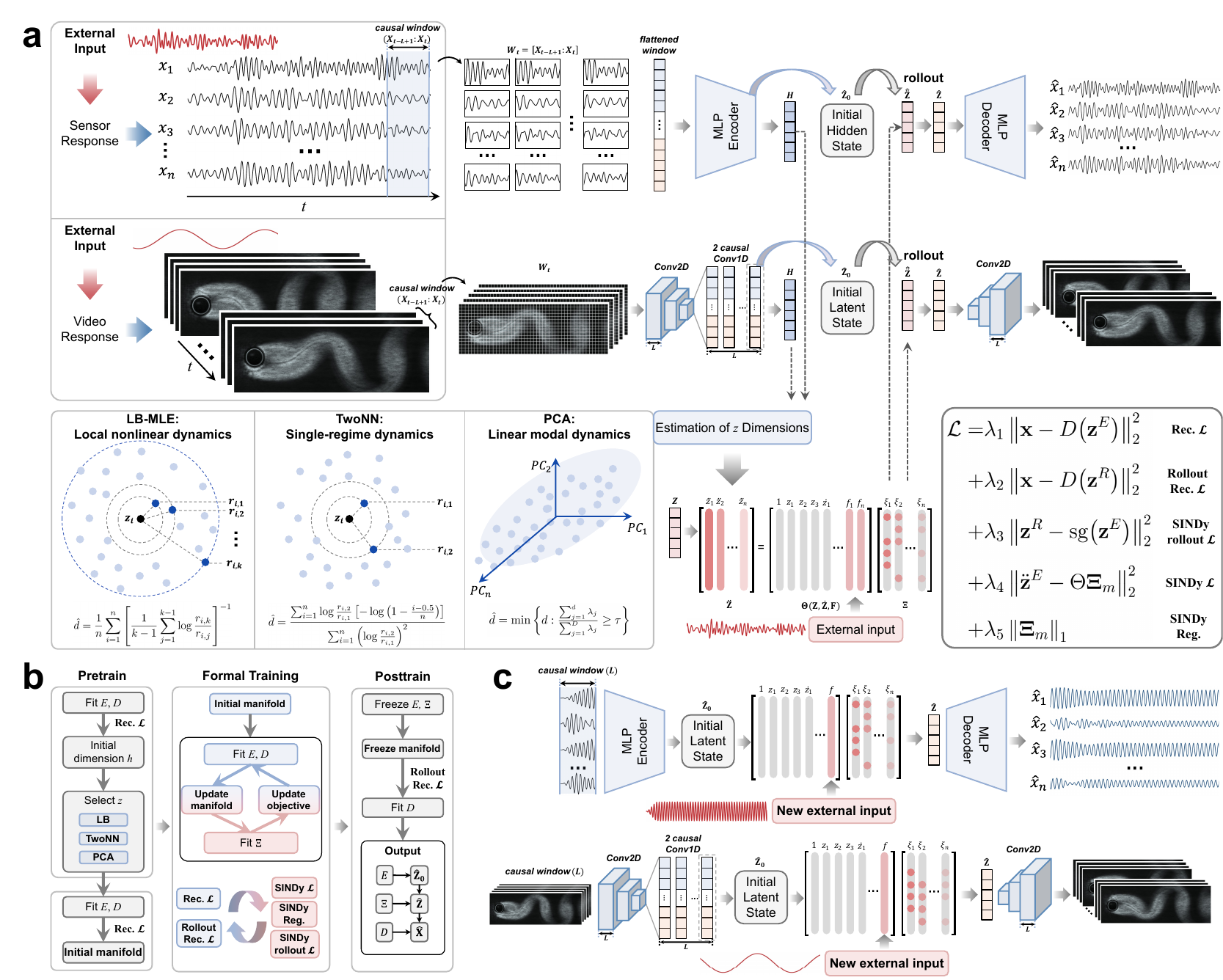}
\caption{\textbf{Architecture, training and offline prediction of FLARE.}
\textbf{a}, FLARE processes either vector-valued sensor responses or video sequences while retaining a separate external-input channel. At each time $t$, a causal response window $\boldsymbol{X}_{t-L+1:t}$ is mapped to a latent state. Vector-valued windows are flattened and processed by a multilayer perceptron (MLP), whereas video windows are encoded spatially by two-dimensional convolutions and aggregated temporally by causal one-dimensional convolutions. The external input is excluded from the encoder and enters only the candidate library $\boldsymbol{\Theta}$ for the latent evolution, whose sparse coefficient matrix $\boldsymbol{\Xi}$ specifies the response equation. Starting from the encoded initial state $\hat{\boldsymbol{z}}_{0}$, the equation is rolled out in latent space and the resulting trajectory is decoded into the full response. The intrinsic dimension of the latent point cloud formed during pre-training is estimated using the Levina--Bickel maximum-likelihood estimator (LB-MLE), the two-nearest-neighbour estimator (TwoNN) or principal component analysis (PCA), according to the geometry of the learned representation. The training objective combines response reconstruction, decoded-rollout reconstruction, latent-rollout consistency, equation consistency and $\ell_{1}$ regularization of the equation coefficients. Here, $E$, $D$ and $\boldsymbol{\Xi}$ denote the encoder, decoder and sparse coefficient matrix, respectively; $L$ denotes the causal-window length; and $\operatorname{sg}$ denotes the stop-gradient operator.
\textbf{b}, FLARE is trained in three stages. Pre-training constructs an initial latent manifold and estimates its dimension. Formal training jointly refines the encoder--decoder representation and the sparse evolution equation by alternating between manifold and equation updates. During post-training, the encoder and identified equation are fixed while the decoder is refined against rollout reconstruction.
\textbf{c}, For offline prediction, the causal response window available at the forecast origin is used only to initialize the latent state. A prescribed new input history then drives the identified latent equation without further access to the true response, and the resulting trajectory is decoded into either vector-valued responses or image sequences.}\label{fig1}
\end{figure}

We introduce the Forced Latent Autoencoder for Response Equations (FLARE), a framework that identifies sparse, input-dependent dynamics while learning a compact representation of high-dimensional responses (Figure~\ref{fig1}). FLARE first estimates the intrinsic dimension of the latent point cloud formed during pre-training, thereby automatically selecting the latent dimensionality required for a compact representation. With this dimensionality determined, an encoder estimates the latent state from a causal time window containing the current and preceding response observations, without access to the external input. The input is instead introduced explicitly into a candidate function library for the latent evolution, whose sparse coefficients specify an explicit response equation. Conditioned on the encoded initial state and a prescribed input history, this equation is integrated in latent space, and the resulting trajectory is mapped by a decoder back to the full response. For offline prediction, FLARE uses only the response history available at and before the forecast origin to initialize the latent state; it then advances the latent equation independently under the prescribed future input and decodes the resulting trajectory into the high-dimensional response. This design separates latent-state estimation from the modelling of external forcing and unifies compact equation discovery with high-dimensional response prediction within a single dynamical framework.

Video sequences constitute an important class of high-dimensional observations in which dynamically relevant variables are not provided as predefined state coordinates but are embedded in pixel fields that evolve over time \cite{Gnesotto2020BrownianMovies,Regazzoni2024LatentDynamicsNetworks}. Discovering equations from such observations therefore requires a compact response representation that preserves both spatial structure and temporal evolution. To extend FLARE to this setting, we replace the fully connected encoder and decoder used for vector-valued responses with convolutional counterparts while retaining the same input-dependent latent equation. A spatial encoder extracts frame-level features, a causal temporal convolution aggregates their recent history into a latent state, and a convolutional decoder reconstructs each predicted frame. The external-input channel and sparse evolution equation remain unchanged, allowing vector-valued measurements and image sequences to be treated under the same modelling principle. This extension enables FLARE to identify compact dynamics from visual responses while predicting their full spatiotemporal evolution. The complete network architectures, latent-dimension selection and training procedures, together with implementation details for vector and video observations, are described in Methods.

\subsection{Recovery of forced dynamics from synthetic high-dimensional observations}\label{subsec2.1}
\begin{figure}[t!]
\centering
\includegraphics[width=\textwidth]{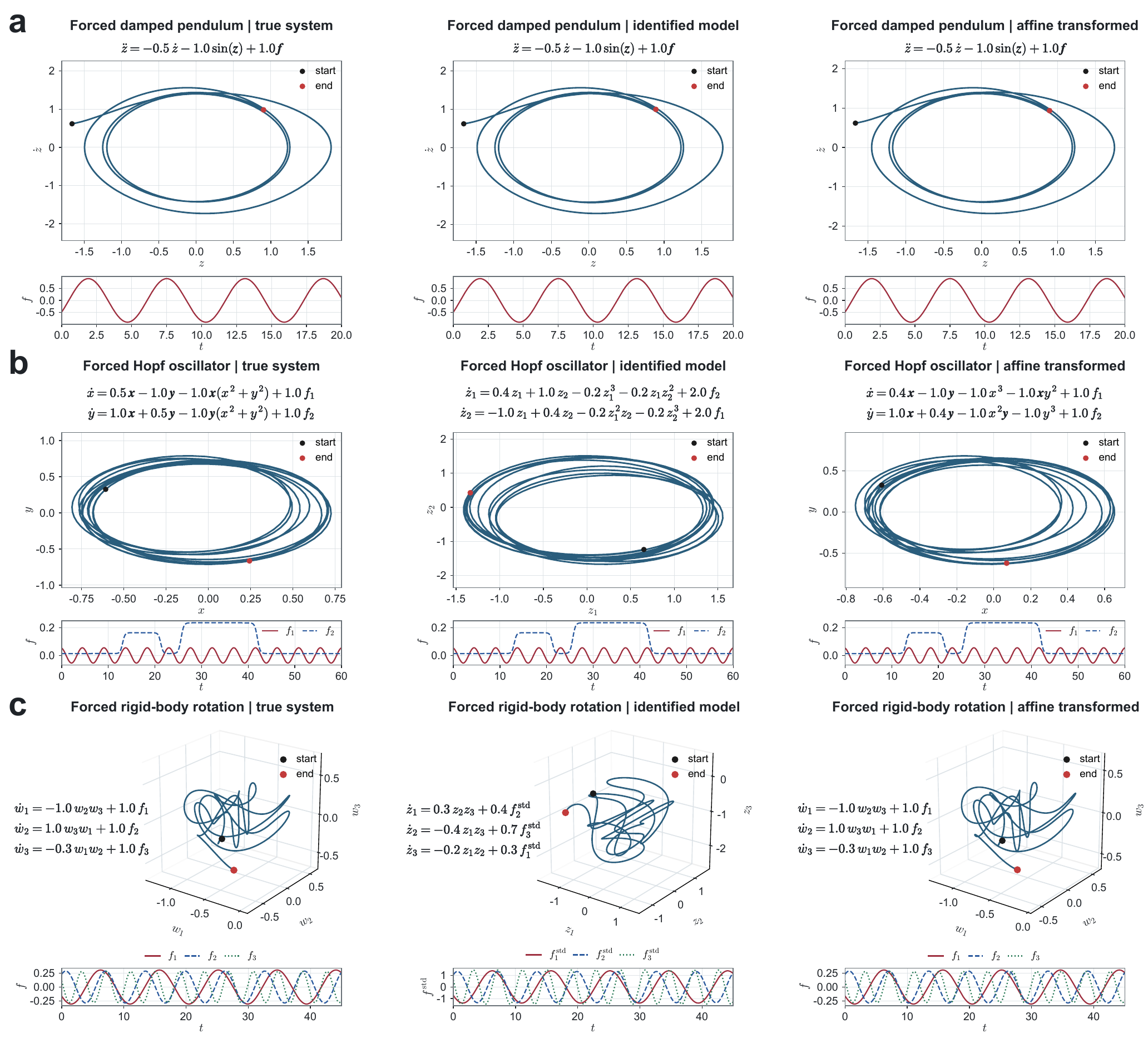}
\caption{\textbf{Recovery of forced dynamics from synthetic high-dimensional observations.}
\textbf{a--c}, Governing equations, phase trajectories and external-input histories for the forced damped pendulum (\textbf{a}), forced Hopf oscillator (\textbf{b}) and forced rigid-body rotation system (\textbf{c}). From left to right, the columns show the ground-truth system in physical coordinates, the model identified directly by FLARE in the learned latent coordinates, and the identified equation after alignment with the physical coordinates.}\label{fig2}
\end{figure}

To test whether FLARE can recover forced dynamics from high-dimensional observations when the ground-truth equations are available, we constructed three synthetic benchmarks with increasing numbers of physical coordinates and input channels: a forced damped pendulum with one generalized coordinate and one input, a two-state forced Hopf normal form with two inputs, and a forced rigid-body rotation system with three states and three inputs. Each system was presented to FLARE only through a 64-dimensional nonlinear observation map:
\begin{equation}
    \boldsymbol{x}(t)
    =
    \boldsymbol{\Phi}\!\left(\boldsymbol{s}(t)\right)
    \in \mathbb{R}^{64},
    \quad
    \boldsymbol{s}(t)\in\mathbb{R}^{d},
    \quad d\in\{1,2,3\}
\end{equation}
where $\boldsymbol{s}$ denotes the physical variables of the underlying system and $\boldsymbol{x}$ denotes the observed high-dimensional response. To examine whether recovery depends on how the low-dimensional dynamics are embedded in the observation space, we adopted three distinct observation maps. The pendulum coordinate was expanded into a set of harmonic components:
\begin{equation}
    \boldsymbol{\Phi}_{\mathrm{trig}}(q)
    =
    \left[
    \sin q,\cos q,\ldots,
    \sin(32q),\cos(32q)
    \right]^{\mathsf{T}}
\end{equation}
whereas the Hopf and rigid-body systems were observed through polynomial mixtures:
\begin{equation}
    \boldsymbol{\Phi}_{\mathrm{poly}}(\boldsymbol{s})
    =
    \boldsymbol{C}\boldsymbol{\psi}(\boldsymbol{s})
    +
    \boldsymbol{b}
\end{equation}
Here, $\boldsymbol{C}$ and $\boldsymbol{b}$ define the channel-specific weights and offsets, while $\boldsymbol{\psi}$ contains monomials up to cubic order for the Hopf system and up to quadratic order for the rescaled rigid-body variables, including their cross terms. For each system, we generated 100 response trajectories under distinct external-input histories and partitioned the complete trajectories into 80 for training, 10 for validation and 10 for testing. The phase trajectories shown in Fig.~\ref{fig2} were generated using input histories from the held-out test set. The resulting benchmarks therefore vary both the intrinsic dimension of the dynamics and the geometry of their high-dimensional observation maps, enabling a controlled assessment of the forced equations identified by FLARE.

Figure~\ref{fig2} compares the ground-truth equations, the equations identified directly by FLARE in the latent coordinates, and those obtained after alignment with the physical coordinates. For each system, the figure also shows the phase trajectories generated by the corresponding equations together with the external-input histories. For the forced damped pendulum, FLARE recovered the governing equation directly from the 64-dimensional trigonometric observations, without any post-processing of the learned coordinate (Fig.~\ref{fig2}a). The identified equation retained the correct linear damping, nonlinear restoring force and dependence on the external input, with coefficients that were nearly identical to their ground-truth values at the reported precision. For the forced Hopf system, the equation identified directly in the learned coordinates did not correspond term by term to the original normal form because the latent variables defined a different coordinate basis. After an invertible affine transformation, the resulting equation approximately recovered the linear growth and rotational terms, the cubic saturation terms and the respective contributions of the two external inputs (Fig.~\ref{fig2}b). The forced rigid-body system required an additional rescaling of the inputs because the external inputs had been standardized during training. We first expressed the identified equation in the original input units and then aligned the latent coordinates through an invertible affine transformation. The transformed equation recovered the three quadratic coupling terms and their corresponding input channels, and reproduced the phase-space evolution of the ground-truth system (Fig.~\ref{fig2}c). Thus, despite increases in the numbers of dynamical variables and input channels and changes in the high-dimensional observation map, FLARE recovered the underlying forced dynamics; when the learned latent coordinate basis did not coincide with the physical variables, an additional coordinate alignment was required.

It is important to note that the need for coordinate alignment reflects the inherent non-uniqueness of latent representations rather than a failure to recover the dynamics. High-dimensional observations do not specify how the axes of a learned latent space should correspond to the underlying physical variables. Without direct supervision from those variables, multiple coordinate systems can represent the same response manifold and describe equivalent dynamics while differing in the orientation, scale, ordering or origin of their coordinates. For the comparisons considered here, the relation between the learned variables $\boldsymbol{z}$ and the physical variables $\boldsymbol{s}$ is expressed as an invertible affine transformation:
\begin{equation}
    \boldsymbol{s}
    =
    \boldsymbol{A}\boldsymbol{z}
    +
    \boldsymbol{b},
    \quad
    \det(\boldsymbol{A})\neq 0
\end{equation}
The latent variables therefore need not correspond component-wise to the physical variables. Consequently, an equation identified in latent coordinates can be dynamically equivalent to the ground-truth system without matching its terms and coefficients before transformation. We therefore evaluate equation recovery up to affine coordinate equivalence, enabling comparison with the physical equations without introducing an unrestricted nonlinear reparameterization. The affine transformations used for all three systems, together with the corresponding equation-conversion procedures, are provided in the Supplementary Information.

Accordingly, FLARE does not require the learned latent coordinates to reproduce the physical variables exactly, but allows them to be equivalent up to an invertible affine transformation. With this coordinate freedom acknowledged, the central criterion for systems without known reduced equations becomes whether the identified dynamics can accurately predict the full high-dimensional response under prescribed inputs. We examine this capability next in three application-scale systems.

\subsection{High-dimensional response prediction in application-scale systems}\label{subsec2.2}
The synthetic benchmarks establish whether FLARE can recover known forced equations from high-dimensional observations. In application-scale systems, however, the effective reduced equations are generally unavailable, precluding a term-by-term comparison with the identified dynamics. The central criterion therefore becomes whether the learned latent equations can generate accurate full-state responses under prescribed external inputs. We examine this capability in three settings: transient heat transfer in a thin plate, which represents spatially distributed diffusive dynamics; the motion of a robotic manipulator, which involves nonlinear coupling among controlled mechanical degrees of freedom; and the seismic response of a structure, which is governed by transient dynamics under non-stationary excitation. Together, these systems encompass distinct dynamical mechanisms, forcing characteristics and response structures, enabling us to assess the applicability of FLARE across different classes of input-driven systems.

\subsubsection{Prediction under held-out input histories}\label{subsubsec2.2.1}
The thin-plate data were generated by solving a two-dimensional heat-conduction model driven by two independently modulated local heaters, with temperature changes at 100 spatially distributed locations forming the observed response. The robotic data were generated by simulating a Franka Emika Panda manipulator in MuJoCo \cite{Todorov2012MuJoCo}; four commanded joint-position histories served as external inputs, while the three-dimensional coordinates of marker points distributed across its seven links formed a 105-dimensional response. The structural data were generated using the two-storey frame benchmark with Bouc--Wen hysteretic links \cite{Vlachas2021BoucWenFrame}, with ground acceleration as the external input and 36 horizontal nodal displacements as the observed response. For each system, complete input--response trajectories were assigned to disjoint training, validation and test sets. The training data were used to learn the model parameters, the validation data to select the model configuration, and the test data exclusively for final evaluation. For each test trajectory, FLARE initialized the latent state using only the response history available at the forecast origin and subsequently advanced the identified equation under the prescribed future input, without access to subsequent response observations. Details of dataset generation, numerical settings, data partitioning and preprocessing are provided in Methods.

\begin{figure}[t!]
\centering
\includegraphics[width=\textwidth]{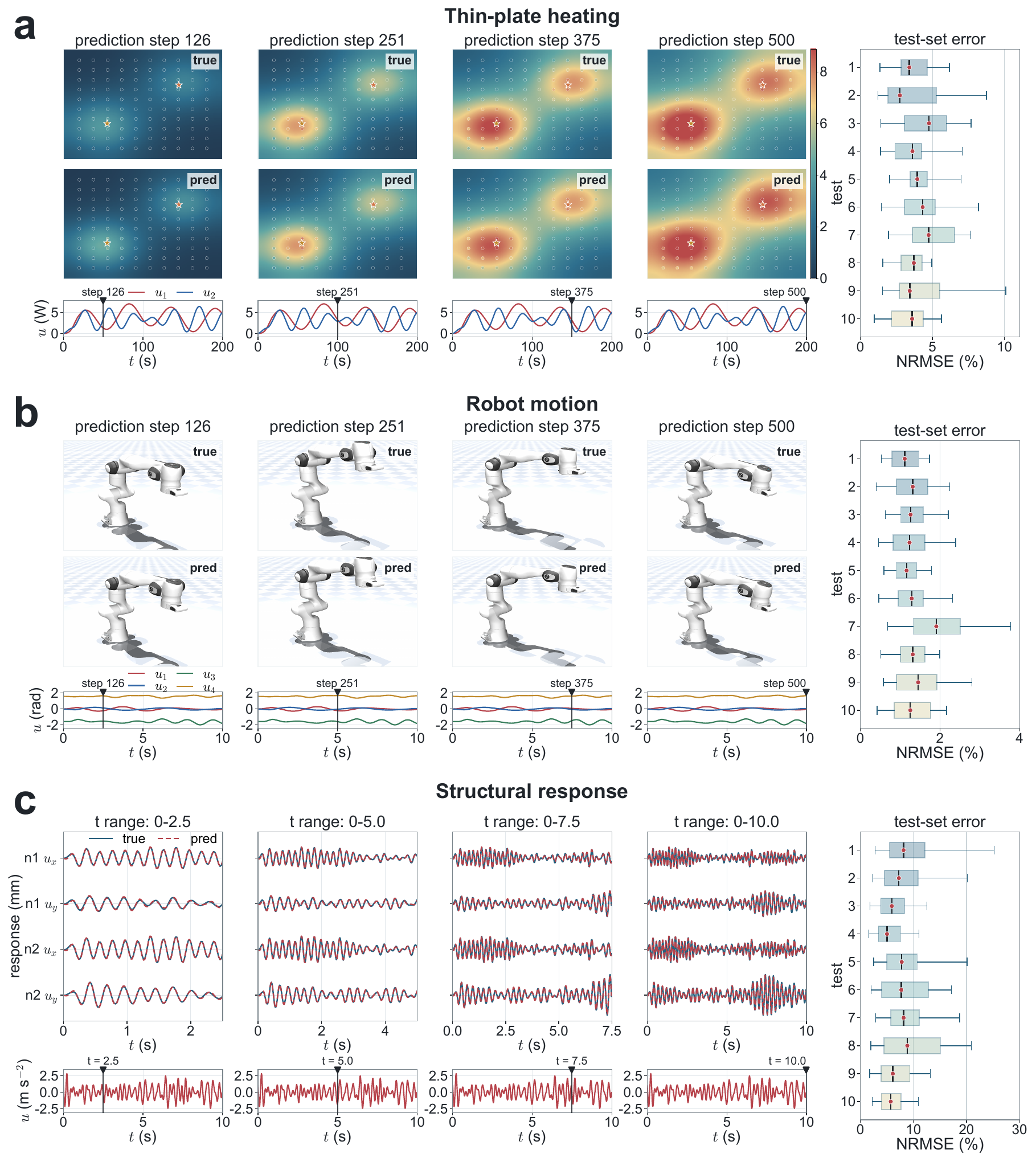}
\caption{\textbf{High-dimensional response prediction in application-scale systems.}
\textbf{a--c}, Offline prediction results for thin-plate heating (\textbf{a}), robotic manipulator motion (\textbf{b}) and structural response under ground excitation (\textbf{c}). In \textbf{a} and \textbf{b}, the first four columns show true and predicted responses for one representative test trajectory at prediction steps 126, 251, 375 and 500, with the corresponding external-input histories shown below. In \textbf{c}, the first four columns show true and predicted displacement histories for representative nodal response channels over the cumulative time ranges corresponding to the same four prediction steps, together with the ground-acceleration input below. The rightmost column shows the NRMSE distributions over the complete 500-step rollouts for all ten test trajectories.}\label{fig3}
\end{figure}

Figure~\ref{fig3} evaluates FLARE on held-out test trajectories from the three application-scale systems. In each case, the latent response equation was initialized from the response history at the forecast origin and then advanced for the full 500-step prediction horizon using only the prescribed future input. For thin-plate heating and robotic manipulator motion, Fig.~\ref{fig3}a,b compare the true and predicted responses for one representative test trajectory at prediction steps 126, 251, 375 and 500; the input histories are shown beneath the response panels, with vertical markers indicating the displayed prediction instants. The rightmost panels summarize the Normalized Root Mean Squared Error (NRMSE) distributions over the complete 500-step rollouts for all ten test trajectories. For structural response, Fig.~\ref{fig3}c reports the same four prediction instants as cumulative time ranges of 0--2.5, 0--5.0, 0--7.5 and 0--10.0 s, comparing predicted and true displacement histories for representative nodal response channels together with the ground-acceleration input. Its rightmost panel likewise summarizes the NRMSE distributions over the complete 500-step rollouts for the ten structural test trajectories. Across these systems, FLARE reproduces the spatial redistribution of the temperature field, the commanded evolution of the manipulator configuration and the phase and amplitude of oscillatory structural responses. Across the ten held-out test trajectories, the median complete-rollout NRMSE was 4.4\%, 1.4\% and 9.9\% for thin-plate heating, robotic manipulator motion and structural response, respectively, showing that the identified latent equations remain predictive over long rollouts and generate accurate high-dimensional responses under held-out input histories.

\subsubsection{Generalization to unseen external inputs}\label{subsubsec2.2.2}
The held-out tests above evaluate prediction for new input--response trajectories whose inputs are of the same type as those used for training. For input-driven systems, a more demanding test is whether a learned response equation can remain predictive when the prescribed input belongs to a forcing regime not covered by the training data. We therefore constructed 10 distinct unseen-input test sets for the three application systems, in which the external inputs differ from the training inputs in their temporal organization, spectral content or coordination across input channels. These changes were designed to probe generalization across forcing conditions while keeping the underlying physical system unchanged. The unseen-input trajectories were not used for training, validation or model selection, and the trained FLARE models were applied to them directly, without retraining or parameter adjustment.

\begin{figure}[t!]
\centering
\includegraphics[width=\textwidth]{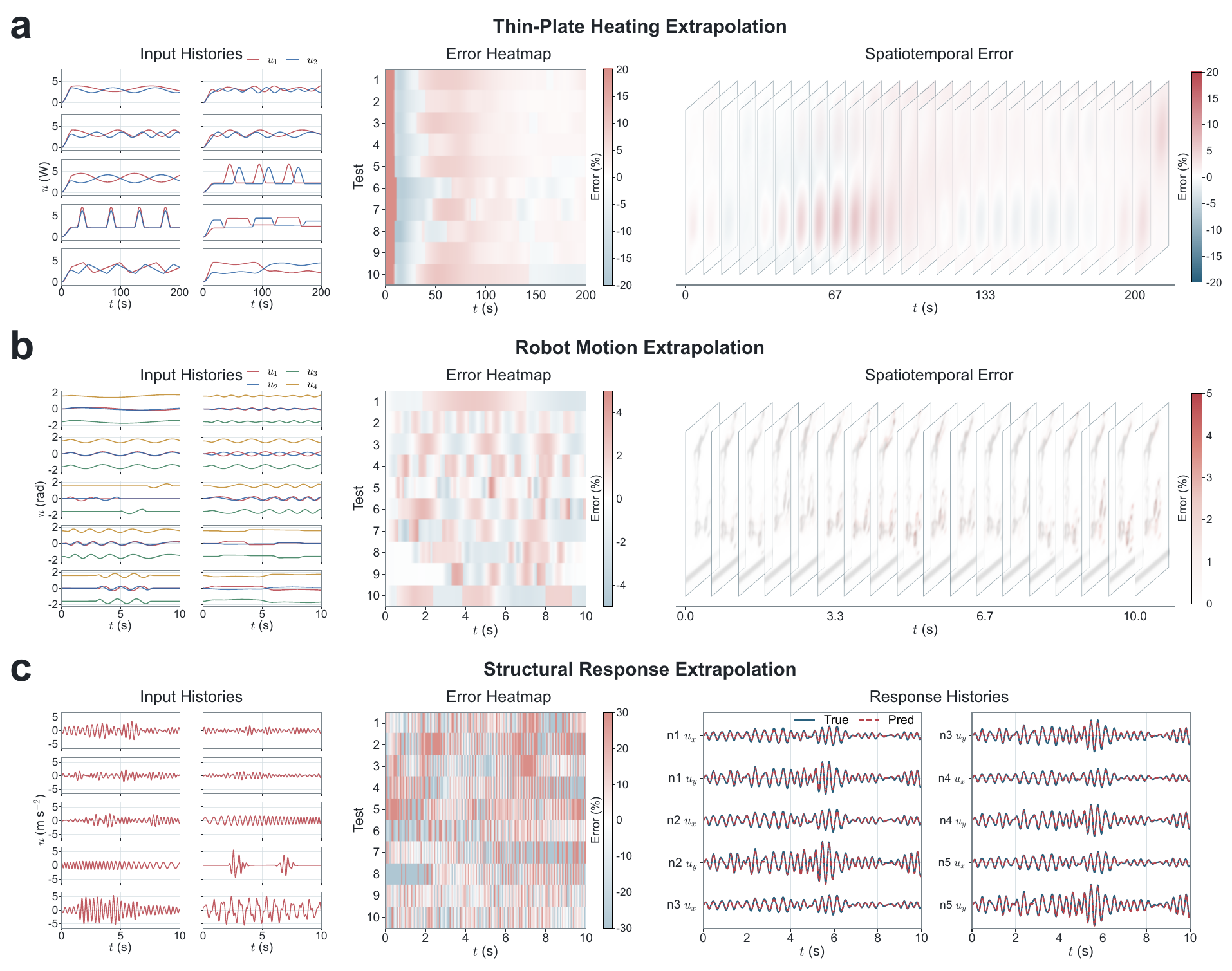}
\caption{\textbf{Generalization to unseen external inputs.}
\textbf{a--c}, Extrapolation results for thin-plate heating (\textbf{a}), robotic manipulator motion (\textbf{b}) and structural response under ground excitation (\textbf{c}). The left column shows the ten unseen external-input histories used for each system. The middle column shows the time-resolved NRMSE over the complete 500-step rollout for the corresponding ten test trajectories. In \textbf{a} and \textbf{b}, the right columns show representative prediction-error snapshots at selected time instants for one test trajectory. In \textbf{c}, the right columns show true and predicted displacement histories for representative nodal response channels from one test trajectory.}\label{fig4}
\end{figure}

Figure~\ref{fig4} evaluates whether the trained FLARE models remain predictive under unseen external inputs. The left column shows the ten unseen-input histories for each system, the middle column reports the corresponding time-resolved NRMSE over the 500-step rollout, and the right column visualizes representative prediction errors or response histories. Across the three tasks, FLARE retained stable predictive behaviour without retraining. In thin-plate heating, the predicted temperature fields followed the redistribution of heat induced by input patterns not present in training, including changes in the relative dominance and temporal organization of the two heaters. In the manipulator system, the predicted marker trajectories preserved the evolving arm configuration under altered joint-command coordination. In the structural system, the predicted displacement histories captured the dominant oscillatory phases and amplitudes produced by unseen ground accelerations. Over the ten unseen-input trajectories, the median complete-rollout NRMSE was 3.0\% for thin-plate heating, 1.6\% for manipulator motion and 13.7\% for structural response. Compared with the held-out tests in Fig.~\ref{fig3}, the error remained comparable for thin-plate heating and manipulator motion, whereas the structural case showed a larger increase under the modified seismic inputs. Nevertheless, the predictions remained bounded and physically consistent over the full 500-step horizons.

We compared FLARE with nine baselines that span the main modelling choices for input-driven response prediction (Table~\ref{tab1}). These include linear and Koopman-based models, represented by DMDc and Deep Koopman with control \cite{Proctor2016DMDc,Shi2022DeepKoopmanControl}; projection- or autoencoder-based sparse equation discovery, represented by POD-SINDYc and AE-SINDYc \cite{Berkooz1993POD,Brunton2016SINDYc,Champion2019CoordinatesEquations}; nonlinear state-space identification, represented by SUBNET \cite{Beintema2023ContinuousTimeSUBNET}; and general-purpose sequence forecasting models, represented by TiDE, TFT, TCN and GRU \cite{Das2023TiDE,Lim2021TFT,Bai2018TCN,Cho2014GRU}. All models used the same training, validation, held-out test and unseen-input trajectories. At prediction time, each model was provided with the same response history at the forecast origin and the same prescribed future input. Performance was evaluated over identical prediction horizons using NRMSE over the first 25\%, 50\% and 100\% of each 500-step rollout, and hyperparameters were selected using only the training and validation data.

On held-out test trajectories, FLARE achieved the lowest full-horizon NRMSE for robotic manipulator motion and structural response, and the second-lowest value for thin-plate heating, where DMDc performed best (Table~\ref{tab1}). Under unseen external inputs, FLARE gave the lowest full-horizon NRMSE in all three systems: 3.20\% for thin-plate heating, 1.56\% for robotic manipulator motion and 14.14\% for structural response. The closest alternatives were TFT for thin-plate heating, Deep Koopman for manipulator motion and GRU for structural response. Several baselines remained competitive over short horizons but degraded over the full rollout, and some models produced divergent predictions in the structural or manipulator tasks. These comparisons indicate that the advantage of FLARE is most pronounced when accurate long-horizon response prediction must be maintained under complex dynamics and unseen forcing conditions.

\begin{table*}[t!]
\centering
\caption{\textbf{Unified comparison with baseline models.}
NRMSE (\%) of FLARE and nine baseline models on held-out test trajectories and unseen-input test trajectories. Errors are reported over the first 25\%, 50\% and 100\% of the 500-step rollout horizon. Lower values indicate better performance. Bold entries denote the lowest finite NRMSE within each dataset and horizon. The symbol $\infty$ indicates that the rollout collapsed or diverged, producing non-finite or unbounded prediction errors.}
\label{tab1}
\small
\setlength{\tabcolsep}{4.2pt}
\begin{tabular}{llrrrrrr}
\toprule
\multirow{2}{*}{Model} &
\multirow{2}{*}{System} &
\multicolumn{3}{c}{Held-out test NRMSE (\%)} &
\multicolumn{3}{c}{Unseen-input test NRMSE (\%)} \\
\cmidrule(lr){3-5} \cmidrule(lr){6-8}
& & 25\% & 50\% & 100\% & 25\% & 50\% & 100\% \\
\midrule
FLARE & Thin plate & 11.26 & 6.97 & 4.38 & 7.07 & 5.94 & \textbf{3.20} \\
      & Manipulator & 1.48 & \textbf{1.49} & \textbf{1.47} & 1.28 & \textbf{1.52} & \textbf{1.56} \\
      & Structure & \textbf{7.09} & \textbf{8.34} & \textbf{9.16} & 17.33 & 13.41 & \textbf{14.14} \\
\midrule
DMDc & Thin plate & 12.19 & 5.78 & \textbf{3.01} & 15.05 & 8.52 & 4.93 \\
     & Manipulator & 1.58 & 21.89 & 18785.22 & 2.20 & 47.02 & 41053.47 \\
     & Structure & $\infty$ & $\infty$ & $\infty$ & $\infty$ & $\infty$ & $\infty$ \\
\midrule
Deep Koopman& Thin plate & \textbf{6.26} & 6.97 & 9.36 & 5.83 & 7.20 & 9.80 \\
             & Manipulator & 1.69 & 1.76 & 1.70 & 1.61 & 1.83 & 1.84 \\
             & Structure & 34.60 & 55.92 & 177.02 & 27.36 & 28.25 & 42.62 \\
\midrule
POD--SINDYc & Thin plate & 24.94 & 21.85 & 15.56 & 20.01 & 16.56 & 9.79 \\
            & Manipulator & 2.66 & 2.68 & 2.65 & 1.89 & 2.32 & 2.37 \\
            & Structure & 116.85 & 117.56 & 115.04 & 167.52 & 159.69 & 140.03 \\
\midrule
AE--SINDYc& Thin plate & 48.25 & 44.96 & 42.24 & 52.90 & 44.19 & 40.84 \\
            & Manipulator & 12.57 & $\infty$ & $\infty$ & 10.71 & $\infty$ & $\infty$ \\
            & Structure & $\infty$ & $\infty$ & $\infty$ & $\infty$ & $\infty$ & $\infty$ \\
\midrule
CT-SUBNET & Thin plate & 6.45 & \textbf{5.59} & 6.60 & \textbf{4.46} & 5.49 & 6.81 \\
          & Manipulator & \textbf{1.21} & 4.04 & 1011.25 & \textbf{1.03} & 4.43 & 2750.94 \\
          & Structure & 22.53 & 128.88 & 894.46 & 27.83 & 218.01 & 1666.23 \\
\midrule
TiDE & Thin plate & 18.61 & 16.14 & 12.77 & 9.24 & 8.16 & 7.63 \\
     & Manipulator & 14.72 & 17.35 & 16.93 & 17.63 & 17.06 & 17.07 \\
     & Structure & 78.67 & 79.44 & 78.02 & 70.81 & 67.45 & 68.02 \\
\midrule
TFT & Thin plate & 8.32 & 6.63 & 5.60 & 5.93 & \textbf{5.40} & 4.48 \\
    & Manipulator & 3.45 & 3.15 & 3.01 & 3.43 & 3.29 & 2.99 \\
    & Structure & 20.04 & 17.32 & 13.90 & 77.39 & 55.80 & 44.33 \\
\midrule
TCN & Thin plate & 7.10 & 8.21 & 9.16 & 6.56 & 12.93 & 14.32 \\
    & Manipulator & 1.92 & 1.90 & 1.85 & 1.86 & 2.00 & 2.04 \\
    & Structure & 12.43 & 13.58 & 14.04 & 18.93 & 15.69 & 15.45 \\
\midrule
GRU & Thin plate & 8.48 & 7.53 & 6.87 & 8.78 & 7.46 & 4.94 \\
    & Manipulator & 1.30 & 2.63 & 3.41 & 2.92 & 3.57 & 4.36 \\
    & Structure & 7.66 & 10.57 & 11.50 & \textbf{11.10} & \textbf{12.14} & 15.00 \\
\bottomrule
\end{tabular}
\end{table*}

Taken together, the three application-scale studies show that FLARE can learn response equations that remain useful beyond the recovery of known synthetic systems. The same framework predicted spatially distributed heat transfer, marker-based manipulator motion and seismic structural responses over 500-step rollouts; it also retained predictive accuracy when the external inputs changed in temporal structure, spectral content or inter-channel coordination. The baseline comparison further indicates that this behaviour is not simply a consequence of using a high-capacity predictor. In regimes where the high-dimensional response is governed by a compact input-driven state, the sparse latent equation provides a stronger inductive structure: it separates state estimation from the representation of forcing, constrains the learned evolution to a low-dimensional dynamical law, and exposes how future inputs enter the response. This structure helps stabilize long rollouts and supports generalization across forcing conditions, while the decoder preserves the connection to the full observed response. The results therefore support the central premise of FLARE: for input-driven systems whose responses are observed in high-dimensional form, explicit latent response equations can provide both accurate full-state prediction and a compact dynamical description of how external inputs shape the response.

\subsection{Extension to video observations}\label{subsec2.3}
Video observations provide a stringent test for response-equation discovery because the measured variables are no longer predefined sensor channels, but pixel fields whose spatial organization evolves in time. In this setting, the effective state must be inferred from visual patterns, while the external forcing remains a separate physical input. FLARE is naturally suited to this extension because its dynamical model is defined in latent space: a response history is encoded into a latent state, a sparse input-dependent equation advances that state, and a decoder maps the resulting trajectory back to the observed response. To adapt FLARE from vector-valued measurements to video sequences, we replaced the fully connected encoder and decoder with convolutional counterparts. A framewise spatial encoder extracts image features, a causal temporal convolution aggregates recent frames into the latent state, and a convolutional decoder reconstructs the predicted frames, while the external-input channel and sparse latent equation remain unchanged.

We evaluated the video extension on three forced systems that differ in both visual structure and underlying dynamics. In the forced Van der Pol oscillator, the low-dimensional oscillator state was rendered as a moving grayscale intensity pattern. In the forced cantilever beam, modal beam deformation was rendered as a grayscale video of the vibrating structure. In the forced cylinder wake, a lattice-Boltzmann flow simulation was used to generate image sequences of the wake field behind a transversely forced cylinder. Each video trajectory was paired with its prescribed scalar input history. The response provided to FLARE consisted of grayscale pixel intensities: the Van der Pol and beam videos had a spatial resolution of $32\times32$ pixels per frame, corresponding to 1,024 response channels, whereas the cylinder-wake videos had a spatial resolution of $32\times96$ pixels per frame, corresponding to 3,072 response channels. Each dataset was partitioned by complete video trajectories into 80 training, 10 validation and 10 test sequences. The Van der Pol and beam datasets contained 400 frames per trajectory, and the cylinder-wake dataset contained 500 frames per trajectory.

\begin{figure}[t!]
\centering
\includegraphics[width=\textwidth]{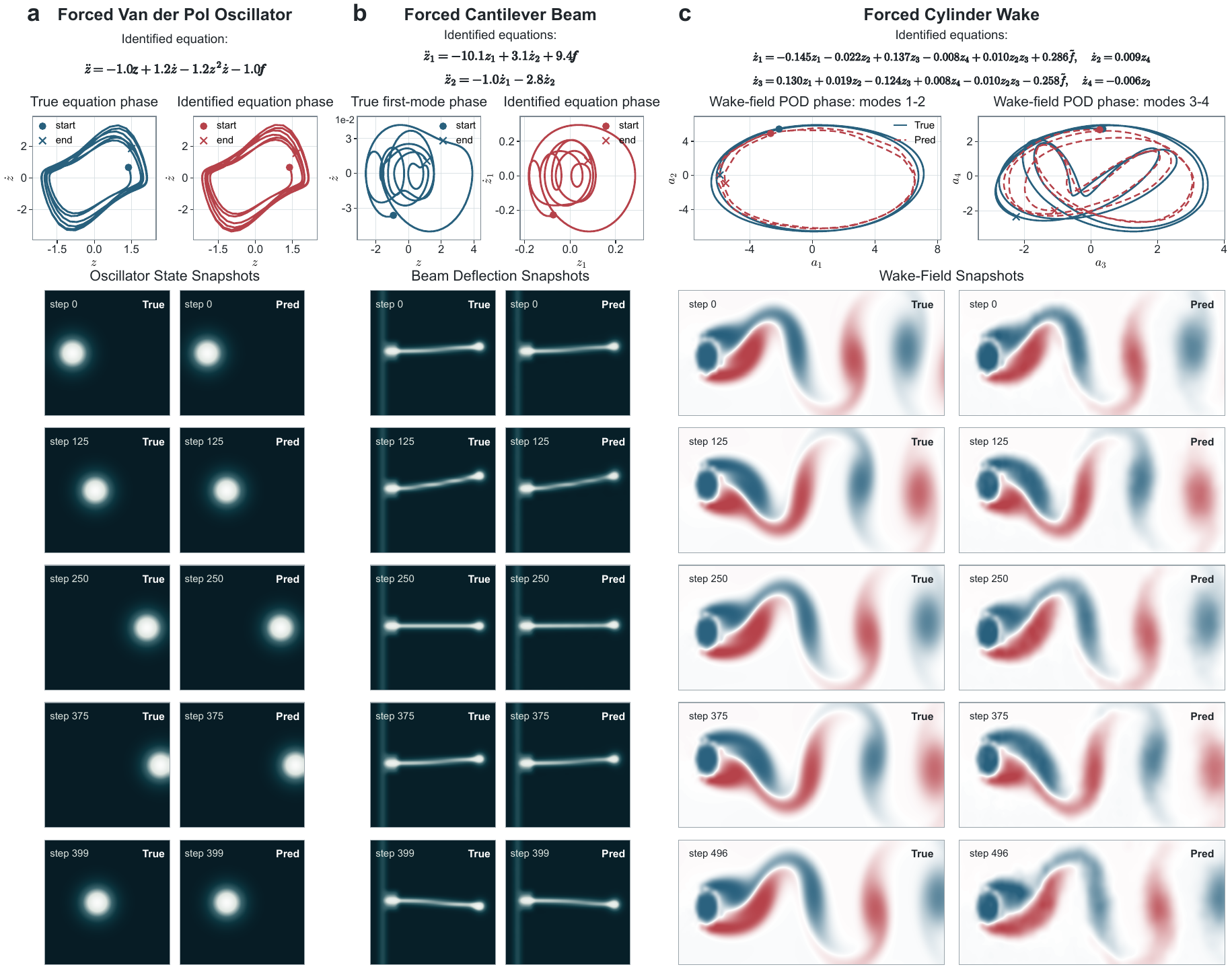}
\caption{\textbf{Extension to video observations.}
\textbf{a--c}, Video prediction and latent-equation results for the forced Van der Pol oscillator (\textbf{a}), forced cantilever beam (\textbf{b}) and forced cylinder wake (\textbf{c}). The top of each panel shows the identified latent equation or equations. In \textbf{a}, the phase portraits compare the trajectory generated by the true oscillator equation with that generated by the identified equation. In \textbf{b}, the phase portraits show the true first-mode phase trajectory and the trajectory generated by the identified latent equation. In \textbf{c}, the phase portraits compare true and predicted wake-field trajectories after projection onto POD modes 1--2 and 3--4. The lower panels show true and predicted video frames from the corresponding test trajectory at selected prediction steps.}\label{fig5}
\end{figure}

The video experiments test a more stringent form of response-equation discovery: the informative variables are not measured as coordinates, but must be inferred from image sequences. Figure~\ref{fig5} therefore evaluates FLARE at three levels: the identified latent equation, the reduced phase-space structure and the decoded pixel-level response. In the forced Van der Pol oscillator, the learned second-order equation recovers the known forced oscillatory dynamics from moving intensity patterns, and its phase portrait remains consistent with the true oscillator trajectory. In the forced cantilever beam, where the model is trained only on grayscale frames and the applied input, the learned latent dynamics organize the video into a phase trajectory consistent with the first bending mode, indicating that the dominant structural coordinate is recovered without prescribing modal amplitudes. In the forced cylinder wake, no closed-form reduced equation is available for direct comparison. We therefore use POD coordinates as an a posteriori diagnostic of the predicted and true wake fields. The agreement of the POD phase portraits shows that FLARE preserves the dominant coherent wake evolution, while the decoded frames further indicate that this reduced dynamics remains coupled to the full vorticity-field images.

\begin{table*}[t]
\centering
\caption{\textbf{Quantitative evaluation of video prediction.}
NRMSE and structural similarity index measure (SSIM) for the three video-observation tasks. Metrics are reported over cumulative prediction windows corresponding to the first 25\%, 50\% and 100\% of the full rollout horizon. Values are averaged over ten held-out test trajectories. Lower NRMSE and higher SSIM indicate better agreement.}
\label{tab2}
\small
\setlength{\tabcolsep}{15pt}
\begin{tabular}{lrrrrrr}
\toprule
\multirow{2}{*}{Video system} &
\multicolumn{3}{c}{NRMSE (\%)} &
\multicolumn{3}{c}{SSIM} \\
\cmidrule(lr){2-4} \cmidrule(lr){5-7}
& 25\% & 50\% & 100\%
& 25\% & 50\% & 100\% \\
\midrule
Forced Van der Pol oscillator & 9.12 & 9.36 & 10.44 & 0.99 & 0.99 & 0.98 \\
Forced cantilever beam        & 11.29 & 14.82 & 17.45 & 0.99 & 0.98 & 0.98 \\
Forced cylinder wake          & 7.52& 8.34& 10.85& 0.93& 0.92& 0.88\\
\bottomrule
\end{tabular}
\end{table*}

Table~\ref{tab2} complements the phase-space and frame-level comparisons with two image-sequence metrics. We report errors over three cumulative prediction windows, denoted by 25\%, 50\% and 100\% of the rollout horizon in the table, corresponding to the first quarter, first half and full predicted video sequence. NRMSE measures the normalized discrepancy of the pixel-valued response, whereas the structural similarity index measure (SSIM) evaluates image similarity in terms of luminance, contrast and spatial structure \cite{Wang2004SSIM}; a value of one indicates identical images. This distinction is important for video dynamics, because small phase shifts of coherent visual structures can increase pointwise error even when the underlying spatial organization remains recognizable. The Van der Pol and beam tests therefore show that FLARE preserves the visual structure of relatively low-dimensional motions over the full rollout, with full-horizon SSIM values of $0.98$. The cylinder wake presents a more stringent case: although SSIM decreases over the rollout, it remains at an acceptable level for a convective flow sequence, indicating that FLARE preserves the principal wake structures while accumulating moderate spatial-phase and pattern-alignment errors over long horizons. Together with the phase-space diagnostics in Fig.~\ref{fig5}, these metrics indicate that the video extension learns a dynamical representation that remains coupled to the pixel field, rather than merely producing plausible individual frames.

These video experiments extend the scope of FLARE beyond tabulated sensor signals and vector-valued fields. When the response is available only as evolving pixels, the framework can still connect visual encoding of the response, input-dependent latent evolution and reconstruction of the full image sequence. The results therefore reinforce the central interpretation of FLARE: its value arises from imposing a dynamical organization on the learned representation. When high-dimensional observations contain a compact controlled state, an explicit latent response equation can serve as an interface between the input, the reduced dynamics and the full spatiotemporal response. Thus, the same modelling principle supports equation discovery and full-response prediction across vector-valued measurements, spatial fields and video observations.

\section{Discussion}\label{sec3}
In this work, we introduced FLARE as a framework for discovering input-dependent response equations from high-dimensional observations. The central difficulty addressed here is that, in many forced systems, neither the effective response coordinates nor the manner in which external inputs act on those coordinates is directly observed; instead, the available data consist of high-dimensional response histories. FLARE approaches this problem by estimating a compact latent state for the response, identifying a sparse evolution law in which the prescribed input enters explicitly, and decoding the resulting latent trajectory back to the full response. In this sense, the learned representation is not used only as a device for reconstruction or forecasting, and the discovered equation is not treated as an isolated low-dimensional model. Rather, representation learning, forced-dynamics identification and high-dimensional response prediction are coupled within a single dynamical framework. This integration enables the model to be initialized from observed response histories, advanced under specified future inputs and assessed through both the structure of the latent equation and the fidelity of the predicted full response.

Explicit latent equations play a role that is distinct from a generic predictive parameterization. When a high-dimensional response admits a compact forced state, an equation in the learned coordinates imposes a dynamical organization on the representation: the coordinates must be sufficient not only to reconstruct observations, but also to close an evolution law driven by measured inputs. This requirement ties the latent state to its future response through a small set of input-dependent terms, making the learned model an object that can be inspected, integrated and evaluated under new forcing histories. The benefit is therefore the possibility of converting a hidden response manifold into a dynamical coordinate system in which forcing, state evolution and full-response reconstruction remain connected. Such structure is particularly useful when the scientific question concerns how changes in input histories alter the ensuing response, in addition to how accurately observed trajectories can be represented.

The experiments support this interpretation at several levels. In the synthetic benchmarks, FLARE recovers the forced dynamics from nonlinear high-dimensional embeddings even when the learned latent coordinates do not coincide with the original physical variables. This observation is important because, in representation learning, equation recovery should not in general be understood as recovery of a prescribed coordinate basis, but rather as recovery of an equivalent dynamical description up to admissible coordinate transformations. The application-scale studies then move beyond settings in which a term-wise ground-truth equation is available, showing that the identified latent response equations can be rolled out over long horizons to generate accurate high-dimensional responses. The unseen-input tests further examine whether the learned input dependence remains predictive outside the input histories used for training and validation, while the video experiments extend the same principle to observations in which the response is encoded in evolving pixel fields. Taken together, these results indicate that FLARE is not tied to a particular measurement format or physical example; its common operating premise is the presence of a learnable compact forced state that remains dynamically coupled to the measured response.

The present study also delineates the scope in which this approach is most naturally applied. FLARE is designed for systems whose high-dimensional responses contain a compact forced state that can be inferred from response histories and advanced under prescribed inputs. Its performance will therefore depend on the informativeness of the observations, the coverage of the input conditions, the selection of the latent dimension and the expressiveness of the candidate library used for the response equation. Coordinate non-uniqueness is also intrinsic to learned representations, so the identified equations should be interpreted in the learned coordinates, or compared with physical variables after an appropriate alignment when such variables are available. These considerations suggest several directions for future work, including uncertainty quantification, adaptive or physics-constrained libraries, incorporation of conservation laws and symmetries, learning from partial or irregular observations, and coupling response-equation discovery with control and experimental design. More broadly, FLARE shows that equation discovery can begin from complex observations without prescribed state variables, while still leading to a model that predicts the full response. In this view, the learned coordinates serve as a dynamical interface between prescribed inputs, reduced evolution and observable system response.

\section{Methods}\label{sec4}
\subsection{FLARE architecture}
FLARE represents an input-driven high-dimensional response through three coupled components: an encoder that estimates a compact latent state from the observed response history, a sparse latent response equation that evolves this state under prescribed external inputs, and a decoder that maps the resulting latent trajectory back to the full response. Let $\boldsymbol{x}_k\in\mathbb{R}^{n_x}$ denote the observed response at time step $k$, and let $\boldsymbol{u}_k\in\mathbb{R}^{n_u}$ denote the corresponding external input. At each time step, the encoder receives a causal response window:
\begin{equation}
    \boldsymbol{W}_k =
    \left(
    \boldsymbol{x}_{k-L+1},
    \boldsymbol{x}_{k-L+2},
    \ldots,
    \boldsymbol{x}_{k}
    \right)
\end{equation}
where $L$ is the history length, and returns a latent state:
\begin{equation}
    \boldsymbol{z}_k = E_{\theta}(\boldsymbol{W}_k)
\end{equation}
The external input is deliberately excluded from this state-estimation step. Instead, it enters the latent evolution model explicitly through the candidate functions used to construct the response equation. This separation makes the latent state a representation of the observed response, while the forcing history acts through the learned dynamics that advance the state.

For vector-valued responses, the window $\boldsymbol{W}_k$ is flattened and processed by a multilayer-perceptron encoder, and the decoder is another multilayer perceptron that maps each latent state to the corresponding high-dimensional response. For video observations, each frame is first processed by a convolutional spatial encoder, and the resulting frame-level features over the causal window are aggregated by a causal temporal convolution to obtain $\boldsymbol{z}_k$. The decoder is replaced by a convolutional decoder that reconstructs the predicted frames. Thus, the architectural components used to process the observation modality differ between vector responses and image sequences, whereas the input-dependent latent equation remains unchanged.

Starting from an encoded initial state $\hat{\boldsymbol{z}}_0=E_{\theta}(\boldsymbol{W}_0)$ and a prescribed future input history, FLARE advances the latent state by integrating the identified response equation:
\begin{equation}
    \hat{\boldsymbol{z}}_{r}
    =
    \mathcal{R}_{\boldsymbol{\Xi}}
    \left(
    \hat{\boldsymbol{z}}_{r-1},
    \boldsymbol{u}_{r-1};
    \Delta t
    \right),
    \quad r=1,\ldots,H
\end{equation}
where $\mathcal{R}_{\boldsymbol{\Xi}}$ denotes one numerical integration step defined by the sparse coefficient matrix $\boldsymbol{\Xi}$, $\Delta t$ is the time step and $H$ is the rollout horizon. The predicted full response is then obtained by decoding the latent rollout:
\begin{equation}
    \hat{\boldsymbol{x}}_{r}=D_{\phi}(\hat{\boldsymbol{z}}_{r})
\end{equation}
This architecture therefore couples state estimation, input-dependent latent evolution and full-response reconstruction in a single model. During rollout, the future response is generated only from the encoded initial state and the prescribed external input, without feeding future observations back into the encoder.

\subsection{Intrinsic latent-dimension estimation}
The latent dimension determines the space in which the response equation is identified. If this dimension is chosen too small, dynamically relevant directions may be merged or discarded; if it is chosen too large, the learned coordinates may contain redundant directions that make the sparse equation less stable and less interpretable. FLARE therefore estimates the intrinsic dimension of the latent point cloud before the formal equation-discovery stage, rather than prescribing the latent dimension solely as a hyperparameter.

In the pre-training stage, an autoencoder with a provisional latent dimension is first trained to reconstruct the response histories. The encoded states from the training trajectories form a point cloud:
\begin{equation}
    \mathcal{Z}
    =
    \left\{
    \boldsymbol{z}_i^{\mathrm{pre}}
    =
    E_{\theta}^{\mathrm{pre}}(\boldsymbol{W}_i)
    \right\}_{i=1}^{N}
\end{equation}
The intrinsic dimension of this point cloud is then estimated and used as the latent dimension in the subsequent FLARE training. We use three complementary estimators because the geometry of the learned point cloud can differ across systems: it may be a locally nonlinear manifold, a point cloud with an approximately single scaling regime, or a nearly linear modal subspace.

For locally sampled nonlinear manifolds, we use the Levina--Bickel maximum-likelihood estimator \cite{Levina2005MLE}. Let $T_j(\boldsymbol{z}_i)$ denote the distance from $\boldsymbol{z}_i$ to its $j$th nearest neighbour. For a neighbourhood size $k$, the local dimension estimate is:
\begin{equation}
    \hat{d}_i(k)
    =
    \left[
    \frac{1}{k-1}
    \sum_{j=1}^{k-1}
    \log
    \frac{T_k(\boldsymbol{z}_i)}
         {T_j(\boldsymbol{z}_i)}
    \right]^{-1}
\end{equation}
A global estimate is obtained by averaging $\hat{d}_i(k)$ over the sampled points and over a range of neighbourhood sizes for which the estimate is stable. This estimator characterizes the local manifold dimension of the latent point cloud from the growth of nearest-neighbour distances across neighbourhood scales.

For latent point clouds whose nearest-neighbour statistics are dominated by a single local scaling regime, we additionally use the two-nearest-neighbour estimator \cite{Facco2017TwoNN}. For each point $\boldsymbol{z}_i$, we compute the ratio between its second- and first-nearest-neighbour distances:
\begin{equation}
    \mu_i
    =
    \frac{T_2(\boldsymbol{z}_i)}
         {T_1(\boldsymbol{z}_i)}
\end{equation}
Here, $T_1(\boldsymbol{z}_i)$ and $T_2(\boldsymbol{z}_i)$ denote the distances from $\boldsymbol{z}_i$ to its first and second nearest neighbours, respectively. Under local uniform sampling on a $d$-dimensional manifold, the cumulative distribution of $\mu_i$ is given by:
\begin{equation}
    F(\mu)
    =
    1-\mu^{-d}
\end{equation}
The dimension is therefore estimated by fitting the empirical nearest-neighbour ratios in the transformed coordinates:
\begin{equation}
    -\log\!\left[1-F_{\mathrm{emp}}(\mu)\right]
    =
    d\log \mu
\end{equation}
The slope of this linear relation gives the intrinsic dimension. This estimator provides a complementary dimension estimate based only on the first- and second-nearest-neighbour distances, reducing the dependence on the choice of a larger neighbourhood size.

When the pre-trained latent point cloud is approximately organized by linear modal directions, principal component analysis is used as a complementary dimension estimator. The centered latent samples are assembled into a matrix $\boldsymbol{Z}_{\mathrm{c}}\in\mathbb{R}^{N\times m}$, where $m$ is the provisional latent dimension used during pre-training. The empirical covariance matrix is then computed as:
\begin{equation}
    \boldsymbol{C}_{\mathcal{Z}}
    =
    \frac{1}{N-1}
    \boldsymbol{Z}_{\mathrm{c}}^{\mathsf{T}}
    \boldsymbol{Z}_{\mathrm{c}}
\end{equation}
Let $\{\lambda_j\}_{j=1}^{m}$ denote the eigenvalue spectrum of $\boldsymbol{C}_{\mathcal{Z}}$, sorted in non-increasing order. The cumulative explained variance of the first $d$ principal directions is defined as:
\begin{equation}
    \rho(d)
    =
    \frac{\sum_{j=1}^{d}\lambda_j}
         {\sum_{j=1}^{m}\lambda_j}
\end{equation}
where $m$ is the provisional latent dimension used during pre-training. The latent dimension is selected as the smallest $d$ for which $\rho(d)$ reaches the prescribed variance threshold. This criterion is used for latent geometries that are close to a linear subspace, where the dominant response directions are well represented by a small number of orthogonal modes.

The final latent dimension is selected from the estimator that best matches the geometry of the pre-trained point cloud. The selected dimension is then fixed for the formal FLARE training, in which the encoder, decoder and sparse response equation are jointly optimized. In this way, the dimensionality of the latent response coordinates is inferred from the data before equation identification, while the subsequent training still determines the coordinates and the input-dependent dynamics jointly.

\subsection{Sparse input-dependent latent response equations}
After the latent dimension has been selected, FLARE identifies an explicit evolution law in the learned response coordinates. Let $\boldsymbol{z}(t)\in\mathbb{R}^{d_z}$ denote the latent response state and let $\boldsymbol{u}(t)\in\mathbb{R}^{d_u}$ denote the prescribed external input. The latent dynamics are represented by a sparse expansion over a library of candidate functions:
\begin{equation}
    \dot{\boldsymbol{z}}(t)
    =
    \boldsymbol{\Theta}
    \left(
    \boldsymbol{z}(t),
    \boldsymbol{u}(t)
    \right)
    \boldsymbol{\Xi}
\end{equation}
Here, $\boldsymbol{\Theta}$ contains candidate terms constructed from the latent coordinates, the external inputs and their interactions, and $\boldsymbol{\Xi}$ is the coefficient matrix to be identified. The library can include constant, linear and polynomial terms of the latent variables, input-dependent terms and state--input coupling terms. The use of a sparse coefficient matrix reflects the assumption that, in an appropriate latent coordinate system, the response evolution is governed by only a small subset of the admissible terms.

In the experiments considered herein, the library variables are chosen according to the form in which the latent response is advanced. For systems represented directly by latent-state evolution, the library is constructed from $\boldsymbol{z}(t)$ and $\boldsymbol{u}(t)$. For oscillatory or mechanical responses, the library may additionally include time-derived latent quantities so that inertial effects can be represented in the learned equation. In all cases, the identified model remains a sparse input-dependent response equation in the learned latent coordinates.

The coefficient matrix $\boldsymbol{\Xi}$ defines the explicit latent response equation. During training, the nonzero entries of $\boldsymbol{\Xi}$ are encouraged by sparsity regularization and sequential pruning to concentrate on a small number of dynamically relevant terms. Once identified, the resulting equation is integrated under a prescribed input history to generate a latent trajectory:
\begin{equation}
    \hat{\boldsymbol{z}}(t)
    =
    \mathcal{I}_{\boldsymbol{\Xi}}
    \left(
    \boldsymbol{z}(t_0),
    \boldsymbol{u}_{[t_0,t]};
    \Delta t
    \right)
\end{equation}
where $\mathcal{I}_{\boldsymbol{\Xi}}$ denotes numerical integration of the sparse response equation from the initial latent state $\boldsymbol{z}(t_0)$, driven by the input history $\boldsymbol{u}_{[t_0,t]}$. The decoded trajectory $D_{\phi}(\hat{\boldsymbol{z}}(t))$ then provides the predicted full response. Thus, the equation coefficients are not fitted only to local derivative information; they are also constrained by whether their integrated trajectories remain consistent with the latent representation and with the reconstructed high-dimensional response.

\subsection{Training objective and optimization}
All experiments were conducted on a workstation equipped with an NVIDIA Tesla V100 GPU. FLARE is trained in three stages: pre-training, formal training and post-training. This staged procedure is used to first construct a non-degenerate response representation, then identify a sparse input-dependent latent equation on this representation, and finally improve the consistency between equation rollouts and high-dimensional reconstructions.

In the pre-training stage, the encoder and decoder are trained as an autoencoder before sparse equation identification. For a response window $\boldsymbol{W}_k$ and its corresponding response $\boldsymbol{x}_k$, the encoded state and reconstructed response are given by:
\begin{equation}
    \boldsymbol{z}_k^{\mathrm{pre}}
    =
    E_{\theta}^{\mathrm{pre}}(\boldsymbol{W}_k),
    \quad
    \tilde{\boldsymbol{x}}_k
    =
    D_{\phi}^{\mathrm{pre}}(\boldsymbol{z}_k^{\mathrm{pre}})
\end{equation}
The pre-training objective is the response reconstruction loss:
\begin{equation}
    \mathcal{L}_{\mathrm{pre}}
    =
    \frac{1}{N}
    \sum_{k=1}^{N}
    \left\|
    \tilde{\boldsymbol{x}}_k-\boldsymbol{x}_k
    \right\|_2^2
\end{equation}
This stage provides an initial latent point cloud from which the intrinsic dimension is estimated. After the latent dimension is selected, the encoder and decoder are reinitialized or reduced to the selected latent dimension for formal FLARE training.

In the formal training stage, the encoder, decoder and sparse response equation are optimized jointly. For a training segment, the encoder first maps each causal response window to an encoded latent trajectory:
\begin{equation}
    \boldsymbol{z}_k
    =
    E_{\theta}(\boldsymbol{W}_k)
\end{equation}
Starting from the encoded initial state $\boldsymbol{z}_0$, the sparse response equation is then integrated under the measured input history to produce a latent rollout:
\begin{equation}
    \hat{\boldsymbol{z}}_{k}
    =
    \mathcal{I}_{\boldsymbol{\Xi}}
    \left(
    \boldsymbol{z}_0,
    \boldsymbol{u}_{[0,k]};
    \Delta t
    \right),
    \quad
    k=1,\ldots,H
\end{equation}
The decoded rollout is obtained as:
\begin{equation}
    \hat{\boldsymbol{x}}_k
    =
    D_{\phi}(\hat{\boldsymbol{z}}_k)
\end{equation}
The formal training objective combines reconstruction, decoded-rollout consistency, latent-rollout consistency, equation consistency and sparsity regularization:
\begin{equation}
    \mathcal{L}_{\mathrm{formal}}
    =
    \lambda_{\mathrm{rec}}\mathcal{L}_{\mathrm{rec}}
    +
    \lambda_{\mathrm{roll}}\mathcal{L}_{\mathrm{roll}}
    +
    \lambda_{\mathrm{lat}}\mathcal{L}_{\mathrm{lat}}
    +
    \lambda_{\mathrm{eq}}\mathcal{L}_{\mathrm{eq}}
    +
    \lambda_{1}\left\|\boldsymbol{\Xi}\right\|_{1}
\end{equation}
The response reconstruction loss preserves the local decoding quality of the learned coordinates:
\begin{equation}
    \mathcal{L}_{\mathrm{rec}}
    =
    \frac{1}{H+1}
    \sum_{k=0}^{H}
    \left\|
    D_{\phi}(\boldsymbol{z}_k)-\boldsymbol{x}_k
    \right\|_2^2
\end{equation}
The decoded-rollout loss requires the trajectory generated by the latent equation to remain consistent with the observed response after decoding:
\begin{equation}
    \mathcal{L}_{\mathrm{roll}}
    =
    \frac{1}{H+1}
    \sum_{k=0}^{H}
    \left\|
    D_{\phi}(\hat{\boldsymbol{z}}_k)-\boldsymbol{x}_k
    \right\|_2^2
\end{equation}
The latent-rollout loss aligns the integrated latent trajectory with the encoded latent trajectory:
\begin{equation}
    \mathcal{L}_{\mathrm{lat}}
    =
    \frac{1}{H+1}
    \sum_{k=0}^{H}
    \left\|
    \hat{\boldsymbol{z}}_k
    -
    \mathrm{sg}\!\left(\boldsymbol{z}_k\right)
    \right\|_2^2
\end{equation}
where $\mathrm{sg}(\cdot)$ denotes the stop-gradient operator. This term constrains the response equation to follow the latent trajectory encoded from the data, while preventing the encoder from being driven toward a trivial latent representation solely to reduce the rollout discrepancy.

The equation-consistency loss fits the sparse library model to the local latent evolution. Let $\boldsymbol{r}_k$ denote the latent quantity to be matched by the response equation, estimated from the encoded trajectory by numerical differentiation or by the corresponding discrete update. The equation-consistency loss is written as:
\begin{equation}
    \mathcal{L}_{\mathrm{eq}}
    =
    \frac{1}{H}
    \sum_{k=0}^{H-1}
    \left\|
    \boldsymbol{r}_k
    -
    \boldsymbol{\Theta}
    \left(
    \boldsymbol{\eta}_k,
    \boldsymbol{u}_k
    \right)
    \boldsymbol{\Xi}
    \right\|_2^2
\end{equation}
Here, $\boldsymbol{\eta}_k$ denotes the latent variables supplied to the candidate library, and $\boldsymbol{\Theta}(\boldsymbol{\eta}_k,\boldsymbol{u}_k)$ is the corresponding input-dependent library. The $\ell_1$ penalty promotes sparsity in $\boldsymbol{\Xi}$, and small coefficients are removed during the sparse-model update so that the final response equation contains only a limited number of active terms.

Several components of the objective are used to avoid degenerate solutions. The reconstruction loss prevents the latent coordinates from losing information about the response. The decoded-rollout loss ensures that the integrated equation remains meaningful in the observation space, rather than only matching local latent derivatives. The stop-gradient form of the latent-rollout loss prevents the encoded trajectory from collapsing toward the equation rollout. Finally, sparsity is introduced together with rollout-based constraints, so that pruning candidate terms does not produce an equation that is locally simple but unstable when integrated.

In the post-training stage, the encoder and sparse response equation are fixed, and only the decoder is refined. The objective is the decoded-rollout reconstruction loss:
\begin{equation}
    \mathcal{L}_{\mathrm{post}}
    =
    \frac{1}{H+1}
    \sum_{k=0}^{H}
    \left\|
    D_{\phi}(\hat{\boldsymbol{z}}_k)-\boldsymbol{x}_k
    \right\|_2^2
\end{equation}
This final stage improves the mapping from equation-generated latent trajectories to the full response without changing the identified latent dynamics. The trained model therefore consists of the encoder used to initialize the latent state, the sparse coefficient matrix defining the response equation, and the decoder used to reconstruct the high-dimensional response from equation rollouts.

\subsection{Offline prediction and evaluation protocol}
All prediction results reported in this work are evaluated under a rollout protocol. At the forecast origin, FLARE receives only the causal response window available up to that time and the prescribed external input history over the prediction horizon. The response window is used to initialize the latent state:
\begin{equation}
    \hat{\boldsymbol{z}}_{0}
    =
    E_{\theta}
    \left(
    \boldsymbol{x}_{-L+1},
    \ldots,
    \boldsymbol{x}_{0}
    \right)
\end{equation}
After this initialization, the future response is not provided to the encoder. The identified latent response equation is advanced using the prescribed future input sequence:
\begin{equation}
    \hat{\boldsymbol{z}}_{k}
    =
    \mathcal{I}_{\boldsymbol{\Xi}}
    \left(
    \hat{\boldsymbol{z}}_{0},
    \boldsymbol{u}_{[0,k]};
    \Delta t
    \right),
    \quad
    k=1,\ldots,H 
\end{equation}
The high-dimensional prediction is then obtained by decoding the latent rollout:
\begin{equation}
    \hat{\boldsymbol{x}}_{k}
    =
    D_{\phi}
    \left(
    \hat{\boldsymbol{z}}_{k}
    \right)
\end{equation}
Thus, the model is evaluated as a forced response generator: it is initialized from past observations and then evolves independently under a specified input history. The same protocol is used for held-out tests, unseen-input tests and video prediction.

For vector-valued responses and spatial fields, prediction accuracy is measured by the normalized root-mean-square error:
\begin{equation}
    \mathrm{NRMSE}
    =
    100
    \times
    \left(
    \frac{
    \sum_{k=1}^{H}
    \left\|
    \hat{\boldsymbol{x}}_{k}
    -
    \boldsymbol{x}_{k}
    \right\|_{2}^{2}
    }{
    \sum_{k=1}^{H}
    \left\|
    \boldsymbol{x}_{k}
    \right\|_{2}^{2}
    }
    \right)^{1/2}
\end{equation}
Here, $\boldsymbol{x}_{k}$ denotes the reference response and $\hat{\boldsymbol{x}}_{k}$ denotes the predicted response. The normalization is performed over the complete evaluated rollout, so the metric measures accumulated prediction error rather than a single-step discrepancy.

For video observations, we additionally report the structural similarity index measure (SSIM) \cite{Wang2004SSIM}. NRMSE measures pixel-level amplitude error, whereas SSIM evaluates similarity in luminance, contrast and spatial structure. For the video tasks, both metrics are computed over cumulative prediction windows corresponding to the first 25\%, 50\% and 100\% of the rollout horizon. The same cumulative-window convention is used to assess how prediction quality changes as the decoded image sequence is advanced farther from the forecast origin.

For baseline comparisons, all models are evaluated with the same response histories, prescribed future inputs, rollout horizons and error metrics. Hyperparameters are selected using only the training and validation sets. If a model produces an unstable rollout whose error becomes unbounded or numerically undefined, the corresponding entry is reported as $\infty$.

\subsection{Experimental protocols}
The experiments were organized to evaluate FLARE at three levels: recovery of known forced dynamics from synthetic high-dimensional observations, response prediction in application-scale systems and extension to video observations. The synthetic benchmarks consisted of a forced damped pendulum, a forced Hopf system and a forced rigid-body system. In each case, trajectories of the low-dimensional forced system were mapped to 64-dimensional observations through nonlinear observation functions, and FLARE was given access only to these high-dimensional responses and the corresponding external inputs. Because the latent coordinates learned by an autoencoder are not required to coincide with the original physical variables, identified equations were compared with the ground-truth systems either directly in the learned coordinates or after an affine coordinate alignment when needed. The full governing equations, observation maps, data-generation parameters and coordinate-alignment procedures are provided in the Supplementary Information.

For the application-scale studies, we considered thin-plate heating, robotic manipulator motion and structural response under ground excitation. These tasks were selected to cover different forms of external forcing, response dimensionality and temporal behaviour. Each dataset was divided into training, validation and held-out test trajectories. In addition, ten distinct sets of unseen-input trajectories was constructed for each application system to evaluate prediction under input histories that were not used for training, validation or model selection. FLARE and all baseline models were evaluated on these unseen-input trajectories without retraining or parameter adjustment.

For the video studies, the response was represented by image sequences. The forced Van der Pol oscillator, forced cantilever beam and forced cylinder wake were used to test whether the same latent-equation framework could operate on pixel-valued observations. The model inputs were the normalized image sequences together with the prescribed external input histories. For the cylinder wake, proper orthogonal decomposition was used only as a posterior diagnostic for comparing the predicted and reference flow-field evolution; it was not supplied to FLARE during training or prediction.

Baseline comparisons were performed under a unified protocol. All models used the same training, validation, held-out test and unseen-input splits, received the same available response histories and prescribed future inputs, and were evaluated over the same rollout horizons with the same metrics. Hyperparameters were selected using only the training and validation data, and the reported errors were computed on test trajectories not used for model selection. Detailed dataset specifications, preprocessing steps, baseline implementations, hyperparameter ranges and additional evaluation results are reported in the Supplementary Information.

\section{Data availability}\label{sec5}
All source data used in this study to reproduce the results are available on GitHub at \url{https://github.com/AI4Slab01/FLARE}.

\section{Code availability}\label{sec6}
All source code used in this study to reproduce the results are available on GitHub at \url{https://github.com/AI4Slab01/FLARE}.

\section*{Acknowledgements}
This work is under the support of National Natural Science Foundation of China (Grant Numbers 52192675 and 51878626).

\section*{Declarations}
The authors declare no conflict of interest.

\bibliography{sn-bibliography}

\end{document}